\def\year{2021}\relax
\documentclass[letterpaper]{article} 
\usepackage{aaai21}  
\usepackage{times}  
\usepackage{helvet} 
\usepackage{courier}  
\usepackage[hyphens]{url}  
\usepackage{graphicx} 
\usepackage{graphicx}  
\usepackage{natbib}  
\usepackage{caption} 
\usepackage{amsmath}
\usepackage{amsfonts}
\usepackage{amssymb}
\usepackage{multirow}
\usepackage{booktabs}
\usepackage{dsfont}
\usepackage{xcolor}

\usepackage{amsthm}

\title{Regularizing Attention Networks for Anomaly Detection in Visual Question Answering}
\author{
    \Large \textbf{Doyup Lee\textsuperscript{\rm 1}, 
                    Yeongjae Cheon\textsuperscript{\rm 2}, 
                    Wook-Shin Han\textsuperscript{\rm 1}\thanks{Corresponding Author}} \\
}

\affiliations{
    POSTECH\textsuperscript{\rm 1}, South Korea \\
    Kakao Brain\textsuperscript{\rm 2}, South Korea \\
    \{doyup.lee, wshan\}@postech.ac.kr\textsuperscript{\rm 1}, yeongjae.cheon@kakaobrain.com\textsuperscript{\rm 2}\\
}

\begin{document}

\maketitle

\begin{abstract}
For stability and reliability of real-world applications, the robustness of DNNs in unimodal tasks has been evaluated.
However, few studies consider abnormal situations that a visual question answering (VQA) model might encounter at test time after deployment in the real-world.
In this study, we evaluate the robustness of state-of-the-art VQA models to five different anomalies, including worst-case scenarios, the most frequent scenarios, and the current limitation of VQA models.
Different from the results in unimodal tasks, the maximum confidence of answers in VQA models cannot detect anomalous inputs, and post-training of the outputs, such as outlier exposure, is ineffective for VQA models.
Thus, we propose an attention-based method, which uses \textit{confidence of reasoning} between input images and questions and shows much more promising results than the previous methods in unimodal tasks.
In addition, we show that a maximum entropy regularization of attention networks can significantly improve the attention-based anomaly detection of the VQA models.
Thanks to the simplicity, attention-based anomaly detection and the regularization are model-agnostic methods, which can be used for various cross-modal attentions in the state-of-the-art VQA models.
The results imply that cross-modal attention in VQA is important to improve not only VQA accuracy, but also the robustness to various anomalies.

\end{abstract}

\section{Introduction} \label{section:intro} 
Visual question answering (VQA) is a challenging task that requires a comprehensive understanding of vision, language, and commonsense knowledge \cite{antol2015vqa,goyal2017making}.
Despite the difficulty, the accuracy of VQA has constantly improved by deep neural networks (DNNs) showing great potential for real-world applications \cite{anderson2018bottom,Kim2018,yu2017multi,yu2018beyond,yu2019deep}.
For example, a VQA system can assist the blind, allowing them to use smartphone to take pictures and pose natural language questions about their images \cite{gurari2018vizwiz}.

Orthogonal to answer accuracy, the capability to recognize abnormal situations is essential for stability and reliability, because there is little control of the test input after deployment of the model in practice.
In the example of blind users, if a VQA model fails to detect anomalous situations and returns wrong answer, then the incorrect answers on abnormal situations will lead to fatal accidents.
However, evaluating robustness of VQA models is only limited to irrelevant questions in previous studies \cite{mahendru2017promise,ray2016question}.

Many studies focus on how DNN classifiers can detect anomalies, such as the unrecognizable \cite{nguyen2015deep}, the irrelevant \cite{ray2016question}, or the out-of-distribution (OOD) inputs \cite{hendrycks2017baseline}.
They commonly calibrate a \textit{predictive confidence} by maximum softmax probability (MSP) in the output predictions \cite{hendrycks2017baseline,liang2018enhancing} and detect OOD inputs.
In addition, \cite{hendrycks2018deep,hein2019relu} use post-training to make the predictions have a uniform distribution on anomalies, and show that the robustness of DNNs is significantly improved.

However, previous studies have focused only on anomaly detection in unimodal tasks such as image or text classification, rather than on tasks with multimodal inputs, such as VQA.
Furthermore, extending anomaly detection to VQA has not been discussed, although it is not trivial and must be carefully conducted because of the bimodality of VQA inputs.
In this study, we categorize various anomalies in VQA into five types according to two criteria: 1) whether the images and/or questions are from OOD or not and 2) whether the pairs of in-distribution (ID) images and questions are answerable by VQA models.
From a distributional perspective, our categorization is a disjoint and complete partition of all possible anomalies in VQA and includes worst-case scenarios, the most frequent scenarios, and the current limitation of VQA models.

Then, we propose a simple attention-based method to calibrate predictive confidences and detect various anomalies in VQA.
We find that MSP, which is the most common in unimodal tasks, can only detect samples with undefined answers, whose answers are not among the answer candidates due to the current limitation of VQA models.
However, MSP cannot detect the worst-case and the most frequent scenarios, which are OOD images/questions and irrelevant pairs of images and questions respectively.
Thus, we use cross-modal attention of VQA models, which associate most related visual objects and question tokens in an input pair.
When an input of VQA models is an anomaly, cross-modal attention networks cannot associate the given image and question, and the anomaly can be detected simply by maximum attention probability (MAP) with low confidence.

To enhance the robustness of VQA models to various anomalies, we also propose a maximum entropy regularization of a cross-modal attention distribution in VQA models.
We find that post-training by outlier exposure \cite{hendrycks2018deep} in unimodal tasks also fails to enhance the robustness of VQA models and causes severe accuracy degradation of a VQA model.
Instead, we show that post-training with a maximum entropy regularization of a cross-modal attention in VQA models can significantly improve anomaly detection by MAP, keeping the accuracy of VQA models.
As the choice of anomalies for post-training is directly related to anomaly detection results \cite{hendrycks2018deep}, we also discuss how to select training anomalies to enhance the robustness of VQA models, considering the bimodality of the inputs and characteristics of VQA.

Our main contributions include:
\begin{itemize}
\item This is the first study to define various anomalies in VQA and evaluate the robustness of recent VQA models to those anomalies. In addition, we show that anomaly detection methods in unimodal tasks cannot be simply generalized in multimodal tasks such as VQA.

\item Our attention-based anomaly detection is technically simple yet powerful. Thanks to the simplicity, our approach is a model-agnostic method, which can be used for various attention modules in the state-of-the-art VQA models. In addition, our maximum entropy regularization of a cross-modal attention distribution can significantly improve the robustness of VQA models and keep the VQA accuracy.

\item We claim that cross-modal attention modules are the key to detecting various anomalies for DNNs with multimodal inputs, including VQA models.
\end{itemize}

\section{The Framework of VQA Models} \label{section:Framework}

\begin{figure}
\centering
\includegraphics[width=7.cm]{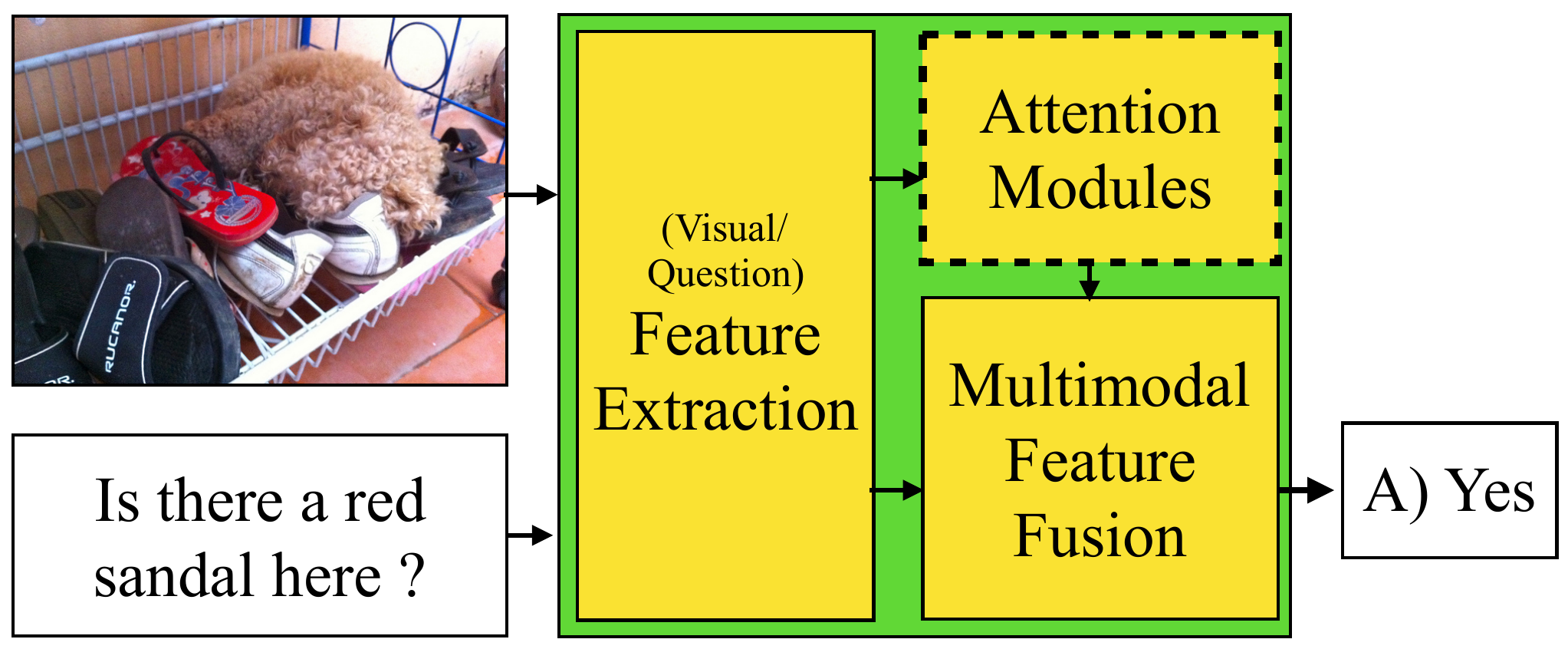}
\captionof{figure}{The framework of VQA models contains attention modules and multimodal feature fusion to predict the answer, given an image and a question.}
\label{VQA_Framework}
\noindent
\end{figure}

A VQA dataset contains a set of triples of answer, image, and question $\mathcal{D}=\left\{(\mathcal{A}, \mathcal{V}, \mathcal{Q}) \right\}$ \cite{antol2015vqa,goyal2017making}. 
A VQA model predicts the answer about a given real-world image and an open-ended question (Fig.~\ref{VQA_Framework}).
The hidden features of $K$ objects (regions) in the image and question (tokens) are extracted by pretrained models \cite{pennington2014glove,ren2015faster,he2016deep}.
Then, the two kinds of features from two modalities are integrated by feature fusion such as element-wise product \cite{anderson2018bottom}, bilinear pooling \cite{fukui2016multimodal}, or multi-modal factorized bilinear (MFB) pooling \cite{yu2017multi}.
Before the integration, attention modules are commonly used to increase the accuracy by cross-modal reasoning between visual objects in the image and the question \cite{anderson2018bottom,yu2018beyond}, or between every pair of visual objects and question tokens \cite{Kim2018,yu2019deep}.
In this paper, we consider VQA models with various types of attention modules.
Finally, the answer is predicted by the joint features of image and question.
The model parameters $\theta$ are trained to maximize expected log likelihood, where $(\mathbf{a},\mathbf{v},\mathbf{q}) \in \mathcal{D}$,
\begin{equation} \label{eq:VQA_train}
\theta^{*}=\underset{\theta}{\operatorname{argmax}}\mathbb{E}_{p_{\mathcal{D}}}\left[\log p_{\theta}\left(\mathbf{a} | \mathbf{v}, \mathbf{q}\right)\right].
\end{equation}

\begin{figure*}
\centering
\includegraphics[width=17cm]{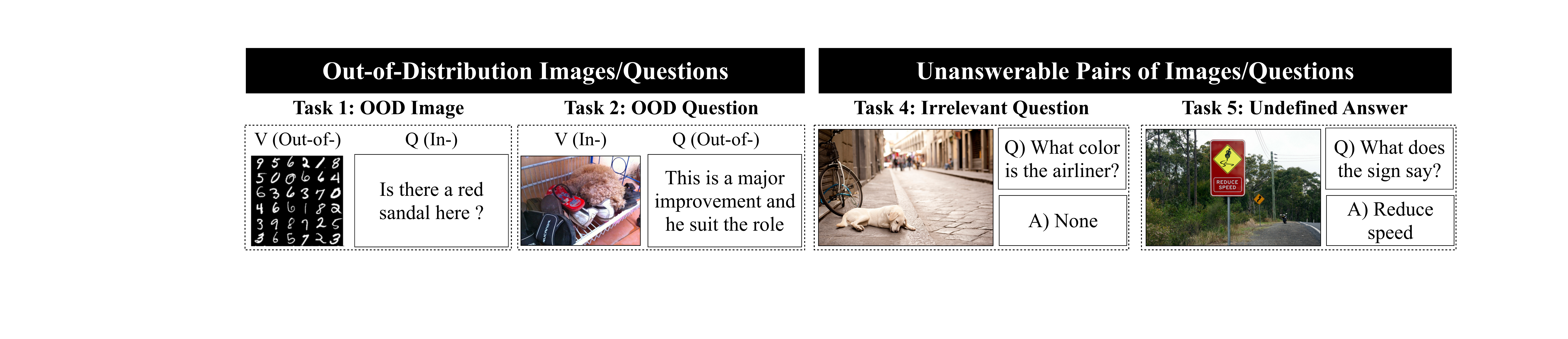}
\caption{Overview of Anomalies in VQA: OOD images and questions, and unanswerable samples with irrelevant questions and undefined answers.}
\label{fig:task_summary}
\end{figure*}

\section{Definition of Anomalies in VQA}\label{section:type}
We define and categorize the five anomaly types in VQA to evaluate the robustness of VQA models.
Considering 1) worst-case scenarios, 2) the most frequent scenarios, and 3) current limitation of VQA models, we divide anomalies in VQA into OOD images/questions and unanswerable pairs of images and questions (irrelevant questions and undefined answers).
Our categorization includes all possible anomalies of $p(\mathbf{a}, \mathbf{v}, \mathbf{q})$ in a distributional approach, and satisfies disjoint and complete partition (Table~\ref{tab:summary}).
Fig.~\ref{fig:task_summary} shows the overview of anomalies in VQA and includes the most extreme case for ease of understanding.
The details of each anomaly are described in the remaining parts.

\subsection{Out-of-distribution Image \& Question}
The typical anomaly is a sample from OOD that differs from training data.
Although OOD samples seem to be unrealistic, worst, and extreme cases in real-world scenarios, detecting them is important because DNNs are not robust but rather over-confident on OOD \cite{hendrycks2017baseline,lee2018simple,hein2019relu}.

\begin{figure}
\centering
\includegraphics[width=5.5cm]{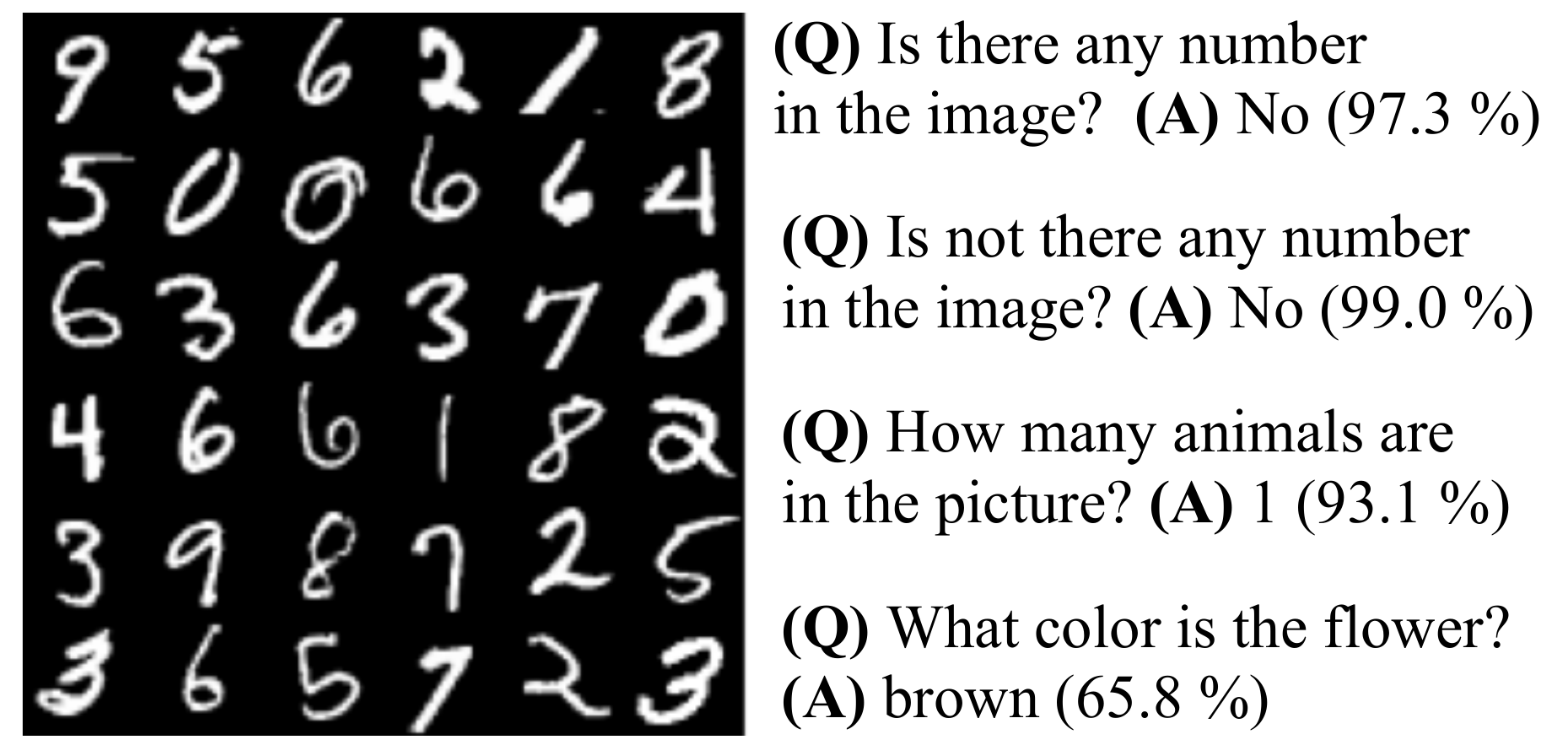}
\captionof{figure}{Examples of unreliable and over-confident misclassification of BUTD for questions about an out-of-distribution MNIST image.}
\label{fig:OOD_I}
\noindent
\end{figure}

\begin{table}
\small
\centering
\begin{tabular}{c|cc|c}
\hline
Task    &   V   &   Q   &   Abnormal Distribution     \\\hline 
1     & OOD & ID & $p(\mathbf{v}_\text{out})$ \\ 
2	&	ID & OOD & $p(\mathbf{q}_\text{out})$ \\
3	&	OOD & OOD & $p(\mathbf{v}_\text{out})$ and $p(\mathbf{q}_\text{out})$\\ 
4	&	ID & ID & $p_\text{out}(\mathbf{v}_\text{in}|\mathbf{q}_\text{in})$ or $p_\text{out}(\mathbf{q}_\text{in}|\mathbf{v}_\text{in})$\\
5	&	ID & ID & $p(\mathbf{a}_\text{out}|\mathbf{v}_\text{in},\mathbf{q}_\text{in})$ \\ \hline
\end{tabular}
\caption{Summary of anomalies in VQA according to ID (in-distribution), OOD (out-of-distribution), and abnormal distribution.}
\label{tab:summary}
\end{table}

\subsubsection{Task 1: Image from Out-of-Distribution}
Task 1 detects the first type of anomalies whose images are from OOD, $p(\mathbf{v}_\text{out})$.
Thus, they are different from images in the original VQA dataset \cite{goyal2017making}.
Then, OOD images can have different visual characteristics, such as different objects, colors, or resolutions.
VQA assumes that an input image contains visual objects in various contexts of the real-world \cite{lin2014microsoft}.
However, VQA models can encounter an OOD image when the image is highly corrupted or selected by users' mistake.

Even though an input image is from OOD but answerable to its question, OOD images need to be detected regardless of the answerability.
In other words, VQA models are susceptible to OOD images and tend to provide arbitrary predictions with high confidence.
We observe that VQA models often predict unreliable and over-confident answers on an OOD image even if it has the correct answer to the question.
For example, the BUTD model \cite{anderson2018bottom} always replies that there is no number in MNIST images regardless of the questions, and the confidences scores are high (Fig.~\ref{fig:OOD_I}).
Including the examples in Fig.~\ref{fig:task_summary} and~\ref{fig:OOD_I}, our experiments also contain more realistic OOD images, such as real-world visual objects with low resolution.

\subsubsection{Task 2: Question from Out-of-distribution}
Task 2 detects the second type of anomalies whose questions are from OOD, $p(\mathbf{q}_\text{out})$.
An OOD question means a non-question sentence without interrogatives.
Questions, such as "Is there a red sandal here?" or "What color is the airliner?", are expected in VQA.
However, after the deployment of the model, VQA models can encounter non-question sentences unconsidered at training time.
When VQA models take a non-question sentence, they have to detect and refuse to answer the input, since there is no right answer to the non-question sentence.
In this task, we evaluate whether a VQA model can distinguish such OOD questions from normal ones.

\subsubsection{Task 3: Image/Question from Out-of-Distribution}
Task 3 detects the third type of anomalies where image and question are both from out-of-distribution, $p(\mathbf{v}_\text{out})$ and $p(\mathbf{q}_\text{out})$.
Although this situation is rare in the real-world, including this task considers extreme cases, making our categorization of anomalies in VQA complete.

\subsection{Unanswerable Pair of Image \& Question}
Although both image and question are from in-distribution, $p(\mathbf{v}_\text{in})$ and $p(\mathbf{q}_\text{in})$, the pair of image and question can be an anomaly, which is unanswerable by a VQA model.
Unanswerable situations occur when the correct answer does not exist because of question irrelevance or the limited capability of the VQA model.
Note that unanswerable pairs are the most frequent and realistic anomalies, because each image and question is similar to training samples.

\subsubsection{Task 4: Irrelevant Question}
Task 4 detects the fourth type of anomalies where each sample has a question irrelevant to the image.
Different from OOD questions, irrelevant questions are sentences with interrogatives.
However, the questions are unrelated to the given input images. 
Out-of-domain questions \cite{kamath2020selective} are also included in this task, because we define out-of-distribution questions as non-question types of sentences.
Although both image and question are from in-distribution, an irrelevant pair of image and question is from out of joint distribution, $p_\text{out}(\mathbf{q}_\text{in}|\mathbf{v}_\text{in})$.

If the image and the question are unrelated to each other, the correct answer requires either external knowledge or does not exist \cite{ray2016question}.
For example, a non-visual question, ``Who is the president of the USA?,'' requires general knowledge irrelevant to the input image.
Moreover, when a question has a visual false-promise, which means that an object implied by the question does not exist in the image, there is no correct answer for the given image and question pair.
In Fig.~\ref{fig:task_summary}, the question asks about an airliner, but no airliner exists in the image.

\subsubsection{Task 5: Undefined Answer}
Task 5 detects the fifth type of anomalies where each sample has an undefined answer, which is not among the answer candidates of a VQA model and is from  $p(\mathbf{a}_\text{out}|\mathbf{v}_\text{in},\mathbf{q}_\text{in})$.
Considering VQA as a prediction task, answer candidates are predefined, and some answers that rarely appear in training data are excluded from answer candidates to improve training efficiency and accuracy \cite{anderson2018bottom}.
Thus, the unanswerability of samples with an undefined answer results not from any abnormality of the input pairs, but from the limited predefined answer candidates. 
The main reasons for rare answers are ambiguous questions, synonyms, and granularity of answers \cite{bhattacharya2019does}, reading numbers or texts.
For example, in Fig.~\ref{fig:task_summary}, the correct answer is ``reduce speed,'' but that answer is not defined in the VQA model because of its rare occurrence.

\section{Anomaly Detection in VQA} \label{section:ad}
In this section, we show how VQA models detect various anomalies without the addition of an extra model or modification of the model architecture.
First, we introduce a confidence-based anomaly detector and its limitation to detect various anomalies in VQA.
Then, we propose the \textit{maximum attention score} as the confidence of reasoning to calibrate the predictive confidence of an input pair of an image and a question. 
How to further classify the types of detected anomalies is interesting future work. 

\subsection{Confidence-based Anomaly Detector} 
A confidence-based anomaly detector $g$ determines an input pair $(\mathbf{v}, \mathbf{q})$ as anomalous if the predictive confidence $S$ is under threshold $\delta$:
\begin{equation} \label{eq:ad_baseline}
    g(\mathbf{v}, \mathbf{q})=\left\{\begin{array}{ll}{1} & {\text{ if  } S({\mathbf{v}, \mathbf{q}}) \leq \delta} \\ {0} & {\text{else }} \end{array}\right.
\end{equation}
To determine the threshold $\delta$ in this anomaly detector, an additional validation dataset can be used in practice.

To compute the confidence $S$ in DNNs, the maximum value of softmax in the output layer (MSP) is commonly used \cite{settles2009active,hendrycks2017baseline}.
\begin{align} 
    S({\mathbf{v}, \mathbf{q}} ; T) &= \max_i{ p_{\theta}(\mathbf{a}_i|\mathbf{v}, \mathbf{q} ; T)} \nonumber \\ 
        &= \max_i{ \frac{\text{exp}(f_i({\mathbf{v}, \mathbf{q}})/T)}{\sum_{j=1}^{N}{\text{exp}(f_j({\mathbf{v}, \mathbf{q}})/T)}}}, \label{eq:conf}
\end{align}
where $f_i$ returns the preactivated output for the $i$-th class in the output layer, $N$ is the number of answer candidates, and $T$ is a temperature parameter.
The temperature is 1.0 in training, and increasing $T$ in test time is known to improve confidence calibration and OOD detection \cite{guo2017calibration,liang2018enhancing}.
Recent studies \cite{liang2018enhancing,hendrycks2018deep} in \emph{unimodal} tasks show that MSP with temperature scaling can detect OOD samples well.
Meanwhile, MSP is still a sensible measure for calibrating predictive uncertainty of VQA models, which are trained with binary cross entropy for multiple correct answers, because the models still use MSP for inference at test time and evaluation of VQA accuracy.

Despite the simplicity and popularity of MSP, we emphasize that MSP fails to detect various anomalies in VQA for two main reasons.
First, MSP is not enough metric to detect whether an input is from abnormal distribution \cite{meinke2019towards}.
MSP does not directly measure $p(\mathbf{v}_\text{in}, \mathbf{q}_\text{in})$, but rather $p(\mathbf{a}_\text{in}|\mathbf{v}_\text{in},\mathbf{q}_\text{in})$.
Thus, MSP can detect a sample with $p(\mathbf{a}_\text{out}|\mathbf{v}_\text{in},\mathbf{q}_\text{in})$.
However, MSP can often fail to detect input pairs of images and questions, which are from abnormal $p(\mathbf{v},\mathbf{q})$, including $p(\mathbf{v}_\text{out})$, $p(\mathbf{q}_\text{out})$, and $p_\text{out}(\mathbf{q}_\text{in}|\mathbf{v}_\text{in})$ (task 1-4).
Second, after the multimodal feature fusion, an abnormal source of a modality vanishes.
For example, although an input image and question are from OOD and ID respectively, the joint features after the feature fusion are hardly distinguishable from those of normal inputs. 

\subsection{Attention-based Anomaly Detection} 
If the joint density of inputs, $p(\mathbf{v},\mathbf{q})$, is explicitly estimated, we can predict the likelihood of an input pair $(\mathbf{v},\mathbf{q})$ and decide whether the pair is from abnormal distribution.
However, the explicit density estimation of multimodal data is computationally expensive and hard to train \cite{Salimans2017PixeCNN,kingma2018glow}.

In this study, we propose attention-based anomaly detection to detect various anomalies from $p(\mathbf{v},\mathbf{q})$.
Instead of using MSP for $S$ in Eq~(\ref{eq:ad_baseline}), we use maximum attention probability (MAP), $A({\mathbf{v}, \mathbf{q}} ; T)$ of a cross-modal attention:
\begin{align} \label{eq:att_score}
    A({\mathbf{v}, \mathbf{q}} ; T) &= \underset{i,j}{\text{max}} A_{ij}(\mathbf{v}, \mathbf{q} ; T)  \nonumber \\
    &= \underset{i,j}{\text{max}} \frac{\text{exp}(a({\mathbf{v}_i, \mathbf{q}_j})/T)}{\sum_{k=1}^{K}\sum_{m=1}^{M}{\text{exp}(a({\mathbf{v}_k, \mathbf{q}_m})/T)}},
\end{align}
where $a$ is a cross-modal attention layer in a VQA model; $A_{ij}$ is the attention score between $i$-th visual object (region) and $j$-th question token; $\mathbf{v}_i$ and $\mathbf{q}_j$ are the features of the $i$-th visual object and $j$-th question token; and $K$ and $M$ are the numbers of visual objects and question tokens.
The temperature parameter is increased only when detecting anomalies, because increasing $T$ affects the prediction results.

We postulate that although MAP does not directly estimate $p(\mathbf{v},\mathbf{q})$, MAP can detect abnormal inputs from $p(\mathbf{v}_\text{out})$, $p(\mathbf{q}_\text{out})$, and $p_\text{out}(\mathbf{q}_\text{in}|\mathbf{v}_\text{in})$.
For example, when the image $\mathbf{v}$ and question $\mathbf{q}$ are both from in-distribution and relevant to each other, we can expect the joint density of the input pair $(\mathbf{v},\mathbf{q})$ to be high.
Together with the high input density, VQA models have high MAP on the input pair, creating a strong attention between a visual object in the image and corresponding question tokens in the question.
In contrast, when either $\mathbf{v}$ or $\mathbf{q}$ is from out-of-distribution, or they are irrelevant, we expect the density of the input pair to be low, and VQA models also have low MAP because they cannot find any strong association between the image and question.

Note that MAP is a model-agnostic metric so it can be used for various attention mechanisms in state-of-the-art VQA models.
If the attention layer does not take all question tokens, but rather uses the context vector of the question \cite{anderson2018bottom,yu2018beyond},
we can notate that $\mathbf{q}_m$ is the context vector and $M=1$ in Eq~(\ref{eq:att_score}).
When a VQA model uses multi-head attentions \cite{Kim2018,yu2019deep}, we use the average of the maximum attention scores in each head over all attention heads.

\subsection{Regularization of Attention Networks for Anomaly Detection}
In unimodal tasks such as image and text classification, post-training of DNNs with known anomalies, such as outlier exposure (OE) \cite{hendrycks2018deep}, has shown remarkable improvement of OOD detection \cite{hendrycks2018deep,hein2019relu}.
Unfortunately, we find that anomaly detection of VQA models does not improve much when we directly exploit OE.

In this section, we introduce how to regularize attention networks by post-training with additional anomalies for boosting anomaly detection of VQA models.
Similar to OE, we \textit{explicitly} fine-tune VQA models to avoid strong attention to anomalies, adding a regularization of attention networks:
\begin{gather} 
\mathbb{E}_{\left(\mathbf{v}, \mathbf{q}\right) \sim P_{\text{in}}}\left[\log p_{\theta}\left(\mathbf{a} | \mathbf{v}, \mathbf{q}\right)\right] \nonumber \nonumber \\
+ \lambda \mathbb{E}_{\left(\mathbf{v}', \mathbf{q}'\right) \sim P_\text{anomaly}}\left[ \sum_{i=1}^{K}\sum_{j=1}^{M} \log\left(1-A_{ij}(\mathbf{v}', \mathbf{q}') \right)  \right] \label{eq:ad_training}
\end{gather}
where $(\mathbf{v}',\mathbf{q}')$ is sampled from selected anomaly datasets, $P_\text{anomaly}$, and $\lambda$ is a hyperparameter.
If the high order attention is used, we also regularize all elements in the multi-order attention maps.

Note that a uniform distribution is the optimal solution for maximizing the regularization term in Eq~(\ref{eq:ad_training}), which is a constraint on $\sum_{i=1}^{K}\sum_{j=1}^{M}{A_{ij}}=1$ such that $A_{ij} \in \left[0,1\right]$. 
Maximizing entropy of the attention distribution makes MAPs on anomalies close to zero, and the VQA models can easily distinguish anomalies from normal samples by the MAPs.

\section{Experiments} \label{section:exp}
\subsection{Experimental Setup}
\subsubsection{VQA Models}
We evaluate four VQA models, which have different attention networks and have shown promising results in recent VQA challenges: BUTD \cite{anderson2018bottom}, MHB+ATT \cite{yu2018beyond}, BAN \cite{Kim2018}, and MCAN \cite{yu2019deep}.

\subsubsection{Datasets}
The VQA v2 dataset \cite{goyal2017making} is used for training and is considered normal.
Test samples of MNIST, SVHN, FashionMNIST, CIFAR-10, and TinyImageNet are used for OOD images.
The 20 Newsgroup, Reuter 52, and IMDB movie review datasets are used for OOD questions.
For irrelevant question datasets, the two test datasets are used: 1) Visual vs. Non-visual Question (VNQ) \cite{ray2016question} contains general knowledge or philosophical questions. 
2) Question Relevance Prediction and Explanation (QRPE) \cite{mahendru2017promise} contains questions with false-premises about the existence of visual objects in the VQA v2 images.
We define answer candidates that occur in the training dataset over nine times, and 4303 samples in the VQA dataset have undefined answers, which occur in the training dataset fewer than nine times.

\subsubsection{Training Setup}
$K=36$ objects are detected by pretrained faster R-CNN \cite{ren2015faster}, and a 2048 dimensional vector for each object is extracted by pretrained ResNet-152 \cite{he2016deep}.
Question tokens are trimmed to a maximum of 14 words, and pretrained GloVe \cite{pennington2014glove} is used for word embedding.
The batch size is 256.

For regularization of the attention network, we use training samples of TinyImage, VNQ, and QRPE for $P_{\text{anomaly}}$ in Eq~(\ref{eq:ad_training}).
Note that there is no overlap of anomaly data between data for training and evaluation.
VQA models can be trained with the regularization from scratch, but we have found that they require a longer training time but have poor accuracy.
For example, a BUTD model has 44 \% VQA accuracy when it is trained with the regularization from scratch.
Thus, we fine-tune the pretrained VQA models in 15 epochs, and the $\lambda$ in Eq~(\ref{eq:ad_training}) is set to 0.00001.
We choose small value of $\lambda$ to balance the magnitude of the original loss and the regularization loss.
At the first epoch of post-training, the regularization loss can have up to 100 times larger value than the original loss, and large $\lambda$ can make the post-training unstable.

\subsubsection{Evaluation}
We fuse the normal and abnormal datasets and evaluate whether VQA models can distinguish anomalies from normal samples.
We use a threshold-free metric, the area under the receiver operating characteristic curve (AUROC), for evaluating OOD and undefined answer detection.
The uninformative detector has 50.0 AUROC.
We use 10 \% of training samples to determine the increased temperature $T$ and $\delta$, maximizing AUROC scores on the samples.

\subsubsection{Compared Methods for Anomaly Detection}
We use the two baselines of anomaly detection for VQA models: the MSP \cite{hendrycks2017baseline} and the maximum attention probability (MAP, ours).
Then, we also compare the AUROCs of the three variants of MSP and MAP with increased temperature (T), outlier exposure (OE), and our regularizing attention networks (RA).
We exclude the results of RA-MSP and OE-MAP, since RA-MAP is significantly better than RA-MSP, and OE-MAP is worse than MAP.

\begin{table}
\small
\centering
\begin{tabular}{l|ccc}
\hline
Accuracy (\%)     & Baseline & OE & Ours \\ \hline 
BUTD	&	62.6 & 54.9(-7.7) & 61.9(-0.5)\\
MHB+ATT	&	63.3 & 62.4(-0.9) & 62.8(-0.5) \\
BAN	&	63.8 & 61.9(-1.9) & 63.7(-0.1) \\
MCAN &  64.3 & 62.0 (-2.3) & 62.4 (-1.9) \\ \hline
\end{tabular}
\caption{VQA Accuracy and its degradation after post-training of VQA models}
\label{tab:acc}
\end{table}

\begin{table*}
\centering
\scalebox{0.95}{
\begin{tabular}{l|c|c|c|c}
\hline
AUROC		 &	BUTD	&	MHB+ATT	&	BAN  & MCAN \\ \hline
Image& \multicolumn{4}{c}{MSP/MSP(T)/OE-MSP(T)/MAP(T)/RA-MAP(T)} \\ \hline 
MNIST	&		60.3/71.5/75.0/89.0/\textbf{97.8}	&	54.2/42.4/\textbf{95.9}/89.9/94.7	&	54.8/35.0/54.1/99.0/\textbf{100}	& 58.7/58.1/64.0/84.1/\textbf{95.1} \\
SVHN	&		60.5/72.8/75.2/90.3/\textbf{97.9}	&	54.1/42.4/\textbf{96.6}/89.7/96.2	&	55.0/35.2/55.5/100/\textbf{100}	&58.8/58.1/64.2/83.6/\textbf{95.2}\\
FashionMNIST	&		60.4/72.2/75.3/89.6/\textbf{97.8}	&	53.9/42.0/\textbf{96.4}/90.5/95.7	&	54.9/35.0/55.4/99.9/\textbf{100} & 58.8/58.1/64.1/84.5/\textbf{95.3}	\\
CIFAR10	&		60.7/73.5/75.5/90.5/\textbf{98.0}	&	54.1/42.3/\textbf{97.1}/89.9/96.9	&	55.0/35.3/56.1/100/\textbf{100}	& 58.7/58.1/64.2/83.5/\textbf{95.3}\\ 
TinyImageNet	&		61.4/75.6/75.5/92.7/\textbf{99.7}	&	53.8/41.6/96.8/91.5/\textbf{99.2}	&	54.8/34.8/59.7/100/\textbf{100}	& 58.9/58.3/64.2/83.4/\textbf{95.1}\\ \hline
Question &  \multicolumn{4}{c}{MSP/MSP(T)/OE-MSP(T)/MAP(T)/RA-MAP(T)} \\ \hline 
20 Newsgroup	&		69.3/79.8/47.1/78.2/\textbf{95.5}	&	54.1/55.0/73.8/78.9/\textbf{92.6}	&	64.0/81.5/62.6/81.7/\textbf{87.3} & 62.3/62.6/73.0/81.1/\textbf{88.7}	\\
Reuters52	&		70.2/81.5/47.5/76.4/\textbf{97.0}	&	50.9/52.0/77.7/77.4/\textbf{94.3}	&	64.3/83.2/60.0/81.7/\textbf{87.3} & 62.0/60.1/75.3/83.9/\textbf{94.2}	\\
IMDB	&		59.9/69.2/45.4/78.2/\textbf{92.8}	&	49.4/50.2/70.0/77.9/\textbf{91.1}	&	56.1/76.3/60.7/78.1/\textbf{82.5} & 57.3/56.3/67.6/85.4/\textbf{90.9}	\\
\hline
\end{tabular}}
\caption{Out-of-distribution detection performance of VQA models.}
\label{tab:v_ood_overall}
\end{table*}

\subsection{Evaluation of VQA Accuracy}
Although post-training for a robust model is known to degrade the accuracy \cite{goodfellow2014explaining,hendrycks2018deep}, we find that OE results in more degradation of VQA accuracy on the VQA v2 validation dataset than our regularization (Table~\ref{tab:acc}).
OE affects all trainable parameters in the VQA models, easily making VQA models unstable, while our regularization affects parameters related to attention networks.
Note that OE severely degrades the accuracy of the BUTD model by 7.7\%.

\subsection{Out-of-Distribution Detection (Task 1-3)}
We analyze the performance of VQA models and anomaly detection methods on various OOD datasets (Table~\ref{tab:v_ood_overall}).
Our experiments include two main results: 1) previous confidence-based approaches (MSP, OE-MSP) fail to detect OOD samples, and 2) our attention-based approaches (MAP, RA-MAP) significantly improve OOD detection in VQA.

\subsubsection{Attention-based Anomaly Detection}

In contrast to the results in unimodal tasks, Table~\ref{tab:v_ood_overall} shows that MSP is not a proper metric for detecting images and questions from out-of-distribution.
Since MSP directly estimates $p(\mathbf{a}|\mathbf{v},\mathbf{q})$, not $p(\mathbf{v},\mathbf{q})$, it fails to detect OOD images and questions.
For example, the AUROCs of MSP (T) in unimodal tasks are close to 100.0 \cite{liang2018enhancing}, but the MSP and MSP(T) of the VQA models are fairly closed to the AUROC of the uninformative detector.
Furthermore, MHB+ATT and BAN rather have more confident predictions on OOD images than normal inputs.
The result is unintuitive, but a similar result, where the OOD samples have higher likelihood than ID samples, is also reported in \cite{choi2018waic,ren2019likelihood}, when ID is more complex than OOD.

Our attention-based anomaly detection (MAP), however, shows superior results to MSP regardless of VQA models.
The AUROCs differ according to VQA models, but all results are promising with AUROCs ($>$ 80.0).
The results show that VQA models do not make a strong attention between images and questions, when they are from OOD.
Furthermore, the promising results mean that instead of explicit estimation of the joint density of $p(\mathbf{v},\mathbf{q})$, MAP can distinguish OOD samples from normal samples.

\subsubsection{The Effect of Regularization of Attention Networks}
OE-MSP in Table~\ref{tab:v_ood_overall} shows that OE fails to improve OOD detection by MSP, in contrast to the results in unimodal tasks \cite{hendrycks2018deep}.
After the multimodal feature fusion in VQA models, a source of abnormality in input images or questions vanishes, and the MSP, which exploit the fused features, can neither detect OOD inputs nor be improved by OE.
Only the OE-MSP(T) of MHB+ATT for Task 1 shows promising results, and we infer the reason from that MHB+ATT has five times larger dimensions of visual features than other VQA models and can remain the abnormality source after the feature fusion.

On the other hand, our maximum entropy regularization of cross-modal attention networks consistently improves the detection of OOD images and questions by MAP.
The results imply that our regularization can be successfully applied in VQA models, allowing them to avoid generating a strong attention when the input image or question is from OOD.
For example, after our regularization, the AUROCs of RA-MAP (T) for all VQA models increase and reach almost perfect OOD detection ($>$ 90.0).

Note that our regularization does not use the OOD datasets, which are used in Table~\ref{tab:v_ood_overall} for testing.
The VQA models can detect all OOD image datasets, although attention networks are regularized by the TinyImageNet training dataset only.
Furthermore, we do not use an OOD question in training, but the robustness of the VQA models is significantly improved by regularizing on irrelevant questions.
We emphasize that OOD datasets of task 1 and 2 are far from VQA tasks.
Nevertheless, MSP and OE fail to detect such easy anomalies, while our attention-based anomaly detection methods can easily detect them.

The results of Task 3 (both OOD image and question) are consistent with Table~\ref{tab:v_ood_overall}. 
We conclude that MSP and OE, which are the most common methods in unimodal tasks, cannot detect OOD images or questions in VQA, but the cross-modal attention with our regularization is the most appropriate to detect unseen OOD samples and improve the capability of the OOD robustness in VQA models.

\subsection{Irrelevant Question Detection (Task 4)}

\begin{table}
\small
\centering
\begin{tabular}{l|cc}
\hline
Accuracy (\%)     & VNQ & QRPE \\ \hline 
Q-Q' SIM \cite{ray2016question}	&	92.3 & ---\\
QPC-Sim \cite{mahendru2017promise}	&	--- & 76.7\\\hline
RA-MAP (BUTD) & \textbf{93.8} & \textbf{78.0} \\
RA-MAP (MHB+ATT) & \textbf{96.4} & \textbf{89.1} \\
RA-MAP (BAN) & 82.0 & 59.7 \\ 
RA-MAP (MCAN) & 72.1 & 56.6 \\ \hline
\end{tabular}
\caption{Comparison of irrelevant question detection models}
\label{tab:irr_q}
\end{table}

In Table~\ref{tab:irr_q}, the attention-based anomaly detection outperforms the previous methods with extra models for irrelevant question detection.
Q-Q' SIM \cite{ray2016question} and QPC-Sim \cite{mahendru2017promise} are the tailored methods, which build extra models, using captioning models \cite{karpathy2015deep} to generate a question relevant to the image and compares it with the input question.
Even though our attention-based anomaly detector does not use additional models to detect irrelevant questions, RA-MAPs (T) of BUTD and MHB+ATT outperform the previous tailored methods.
Moreover, our method can also be applied to detect other types of anomalies, including irrelevant questions.

Compared to BUTD and MHB+ATT, BAN and MCAN have a room for improvement of irrelevant question detection.
BAN and MCAN use the pairwise relationship between all question tokens and visual objects in their cross-modal attention networks, along with multiple heads of attention.
Thus, one of the attention heads might pay strong attention to interrogatives in irrelevant questions.
In this study, we focus on the importance of cross-modal attention for anomaly detection in VQA.

\begin{table}
\small
\centering
\begin{tabular}{l|cccc}
\hline
MSP/MAP & BUTD         & MHB+ATT     & BAN & MCAN         \\ \hline
AUROC            & 87.2/51.5  & 90.7/51.3 & 85.3/55.5 & 81.3/71.5 \\
\hline
\end{tabular}
\caption{Undefined answers detection results}
\label{tab:undefined}
\end{table}

\subsection{Undefined Answer Detection (Task 5)}
Although MSP cannot detect OOD images and questions, and irrelevant questions, Table~\ref{tab:undefined} shows that for detecting samples with undefined answers, MSP achieves higher accuracy than MAP.
MSP directly estimates $p(\mathbf{a}_\text{in}|\mathbf{v}_\text{in},\mathbf{q}_\text{in})$ and has low value on a sample with undefined answers from $p(\mathbf{a}_\text{out}|\mathbf{v}_\text{in},\mathbf{q}_\text{in})$.
Thus, MSP can successfully detect samples with undefined answers, but is limited to detect them.

MAP cannot detect samples with undefined answers, because the images and questions are not from abnormal $p(\mathbf{v},\mathbf{q})$.
Although the correct answer is undefined among the answer candidates, there exists the correct answer between the input image and question.
Then, as in the case of normal samples, VQA models can generate proper attention between the correlated visual object and the word token as a confident reasoning of the answer.

\begin{table}
\small
\centering
\begin{tabular}{l|cccc}
\hline
AUROC & CIFAR10         & Reuters52     & QRPE (test) \\ \hline
BUTD (MAP)            & 90.5  & 76.4 & 49.6\\ \hline
+Tiny     & 99.9  & 79.0 & 47.1 \\
+IMDB     & 57.6  & 99.8 & 46.9 \\
+Tiny, IMDB          & 99.8 & 99.9 & 44.3  \\
+Tiny, QRPE         & 97.8  & 87.8 & 89.3\\
+Tiny, VNQ, QRPE& 98.0 & 97.0 & 84.8 \\
\hline
\end{tabular}
\caption{AUROCs of BUTD for detecting CIFAR10, Reuters52, and QRPE datasets. TinyImageNet, IMDB, VNQ, and QRPE datasets are used for our regularization}
\label{tab:selection_anomalies}
\end{table}

\subsection{Ablation Study}
\subsubsection{Selection of Anomaly Datasets for Regularization}
The selection of abnormal datasets for the post-training, $P_\text{anomaly}$ in Eq~(\ref{eq:ad_training}), is important because considering all possible anomalies at training time is impossible.
Thus, we compare the performance at detecting OOD images (CIFAR10), questions (Reuters52), and irrelevant questions (QRPE) ,according to the change of selection of $P_\text{anomaly}$ (Table~\ref{tab:selection_anomalies}).

Using anomalies of only a certain modality for the regularization of VQA models does not improve detection of anomalies in the other modality.
Anomalies in one modality do not affect the encoder of another modality at the post-training.
For example, when we use only OOD images (Tiny) or questions (IMDB) for the regularization, unseen OOD images (CIFAR10) or questions (Reuters52) are well detected respectively.
However, the detection of abnormal inputs in the other modality is not improved.
Thus, regularizing both modalities is necessary for the robustness of VQA models to anomalies from both modalities.

For selecting abnormal questions, irrelevant questions allow VQA models to detect both OOD and irrelevant questions.
When IMDB sentences are selected instead of irrelevant questions, the regularization cannot remove the unconditional bias of attention networks on interrogatives regardless of the relevance of an input question and an image.
However, after regularizing with irrelevant questions (VNQ, QRPE), the model also detects OOD questions (Reuters52) because OOD questions are much easier to detect than irrelevant questions.
Note that OOD questions contain no interrogatives and are also irrelevant to the input images.

\subsubsection{Scope of Post-Training for Outlier Exposure}
OE has shown severe degradation of VQA accuracy, and the scope of trainable parameters in post-training is related to unstable performance.
In post-training, OE updates all trainable parameters of VQA models to predict uniform scores over answer candidates.
Thus, OE can make attention modules unable to associate a question with visual objects and severely degrade the VQA accuracy (Table~\ref{tab:acc}).
For example, when OE updates parameters only after the feature fusion, the accuracy drop of BUTD has halved from  $-7.7$ \% to $-3.7$ \%, and the anomaly detection performance remains the same as in the paper.

\section{Related Work} \label{section:related_work}
MSP-based OOD detection has mainly been studied, and it shows promising results for unimodal tasks.
The MSP is a simple yet powerful method for OOD detection, when temperature scaling or input preprocessing is combined \cite{hendrycks2017baseline,liang2018enhancing}.
Moreover, \cite{hendrycks2018deep,hein2019relu} use the post-training of DNNS to predict uniform distribution on abnormal samples and enhance MSP to detect unseen OOD samples almost perfectly.
Meanwhile, \cite{meinke2019towards} show that MSP may not be a metric enough to detect OOD inputs.
Our study is the first on OOD detection in multimodal tasks such as VQA, and shows that MSP cannot detect OOD images and questions, or irrelevant questions.

Few studies consider abnormal situations in VQA, but are confined to limited tasks.
\cite{bhattacharya2019does} investigate why annotators provide different answers to the same visual question.
\cite{mahendru2017promise,ray2016question} mainly cover detection of irrelevant questions, but to quantify question relevance, they build an extra tailored model to generate a question relevant to the image and compare the input question with the generated questions.
We define anomaly detection in VQA more generally and show how VQA models can detect irrelevant questions by attention networks without any extra or tailored model.

Some studies regularize attention weight distribution for various purposes.
In machine translation, abstractive summarization, and query-driven multi-instance learning, the attention distribution is regularized to be sharp or uniform to increase their performance \cite{zhang2018attention,hsu2020query}.
In this study, we regularize attention networks to improve the robustness of VQA models to various anomalies.

\section{Conclusions} \label{section:conclusion}
For a VQA system to be safe in the real-world, the models have to be generalized on unseen abnormal samples, having low predictive confidence.
We have defined the five anomaly types in VQA according to out-of-distribution and answerability, and have evaluated the robustness of four VQA models to defined anomalies.
In contrast to the major results in unimodal classification, we find that MSP and OE are limited to detecting various anomalies from $p(\mathbf{v},\mathbf{q})$

In this study, we propose the attention-based method and regularization of attention networks to significantly improve anomaly detection of VQA models.
Cross-modal reasoning (i.e., attention) improves not only VQA accuracy, but also the robustness to various abnormal situations in VQA.
Our method also conserves the VQA accuracy; detects OOD images and questions almost perfectly; and achieves a new state-of-the-art detection for irrelevant questions.

In future work, we believe that further classification of anomalies will offer promise for distinguishing various anomalies.
Furthermore, an analysis of anomaly detection on a range of distributional shifts would be an interesting future work.
Meanwhile, elaborating attention-based anomaly detection for pairwise and multiple heads attentions is worth exploration to improve irrelevant question detection.
We have observed that VQA accuracy is easily degraded in post-training when the VQA model contains many attention heads. Thus, finding an optimal architecture with multi-head attention for accurate and robust VQA models would be an interesting future work.
Moreover, user studies of anomaly detection in VQA for real-life scenarios would also be an interesting future work.

\section{Acknowledgements}
The authors are grateful to Minsu Cho at POSTECH and the reviewers for their feedback and insightful discussions.
This work was supported by Institute of Information communications Technology Planning Evaluation(IITP) grant funded by the Korea government(MSIT) (No. 2018-0-01398, Development of a Conversational, Self-tuning DBMS).

{
\bibliography{main}

\begin{thebibliography}{38}
\providecommand{\natexlab}[1]{#1}
\providecommand{\url}[1]{\texttt{#1}}
\providecommand{\urlprefix}{URL }
\expandafter\ifx\csname urlstyle\endcsname\relax
  \providecommand{\doi}[1]{doi:\discretionary{}{}{}#1}\else
  \providecommand{\doi}{doi:\discretionary{}{}{}\begingroup
  \urlstyle{rm}\Url}\fi

\bibitem[{Anderson et~al.(2018)Anderson, He, Buehler, Teney, Johnson, Gould,
  and Zhang}]{anderson2018bottom}
Anderson, P.; He, X.; Buehler, C.; Teney, D.; Johnson, M.; Gould, S.; and
  Zhang, L. 2018.
\newblock Bottom-up and top-down attention for image captioning and visual
  question answering.
\newblock In \emph{Proceedings of the IEEE Conference on Computer Vision and
  Pattern Recognition}, 6077--6086.

\bibitem[{Antol et~al.(2015)Antol, Agrawal, Lu, Mitchell, Batra,
  Lawrence~Zitnick, and Parikh}]{antol2015vqa}
Antol, S.; Agrawal, A.; Lu, J.; Mitchell, M.; Batra, D.; Lawrence~Zitnick, C.;
  and Parikh, D. 2015.
\newblock Vqa: Visual question answering.
\newblock In \emph{Proceedings of the IEEE international conference on computer
  vision}, 2425--2433.

\bibitem[{Bai et~al.(2018)Bai, Fu, Zhao, and Mei}]{bai2018deep}
Bai, Y.; Fu, J.; Zhao, T.; and Mei, T. 2018.
\newblock Deep attention neural tensor network for visual question answering.
\newblock In \emph{Proceedings of the European Conference on Computer Vision
  (ECCV)}, 20--35.

\bibitem[{BeSpecular(2020)}]{bespecular:2020}
BeSpecular. 2020.
\newblock \url{https://www.bespecular.com}.

\bibitem[{Bhattacharya, Li, and Gurari(2019)}]{bhattacharya2019does}
Bhattacharya, N.; Li, Q.; and Gurari, D. 2019.
\newblock Why Does a Visual Question Have Different Answers?
\newblock In \emph{Proceedings of the IEEE International Conference on Computer
  Vision}, 4271--4280.

\bibitem[{Choi, Jang, and Alemi(2018)}]{choi2018waic}
Choi, H.; Jang, E.; and Alemi, A.~A. 2018.
\newblock Waic, but why? generative ensembles for robust anomaly detection.
\newblock \emph{arXiv preprint arXiv:1810.01392} .

\bibitem[{Fukui et~al.(2016)Fukui, Park, Yang, Rohrbach, Darrell, and
  Rohrbach}]{fukui2016multimodal}
Fukui, A.; Park, D.~H.; Yang, D.; Rohrbach, A.; Darrell, T.; and Rohrbach, M.
  2016.
\newblock Multimodal compact bilinear pooling for visual question answering and
  visual grounding.
\newblock \emph{arXiv preprint arXiv:1606.01847} .

\bibitem[{Goodfellow, Shlens, and Szegedy(2014)}]{goodfellow2014explaining}
Goodfellow, I.~J.; Shlens, J.; and Szegedy, C. 2014.
\newblock Explaining and harnessing adversarial examples.
\newblock \emph{arXiv preprint arXiv:1412.6572} .

\bibitem[{Goyal et~al.(2017)Goyal, Khot, Summers-Stay, Batra, and
  Parikh}]{goyal2017making}
Goyal, Y.; Khot, T.; Summers-Stay, D.; Batra, D.; and Parikh, D. 2017.
\newblock Making the V in VQA matter: Elevating the role of image understanding
  in Visual Question Answering.
\newblock In \emph{Proceedings of the IEEE Conference on Computer Vision and
  Pattern Recognition}, 6904--6913.

\bibitem[{Guo et~al.(2017)Guo, Pleiss, Sun, and
  Weinberger}]{guo2017calibration}
Guo, C.; Pleiss, G.; Sun, Y.; and Weinberger, K.~Q. 2017.
\newblock On calibration of modern neural networks.
\newblock In \emph{Proceedings of the 34th International Conference on Machine
  Learning-Volume 70}, 1321--1330. JMLR. org.

\bibitem[{Gurari et~al.(2018)Gurari, Li, Stangl, Guo, Lin, Grauman, Luo, and
  Bigham}]{gurari2018vizwiz}
Gurari, D.; Li, Q.; Stangl, A.~J.; Guo, A.; Lin, C.; Grauman, K.; Luo, J.; and
  Bigham, J.~P. 2018.
\newblock Vizwiz grand challenge: Answering visual questions from blind people.
\newblock In \emph{Proceedings of the IEEE Conference on Computer Vision and
  Pattern Recognition}, 3608--3617.

\bibitem[{He et~al.(2016)He, Zhang, Ren, and Sun}]{he2016deep}
He, K.; Zhang, X.; Ren, S.; and Sun, J. 2016.
\newblock Deep residual learning for image recognition.
\newblock In \emph{Proceedings of the IEEE conference on computer vision and
  pattern recognition}, 770--778.

\bibitem[{Hein, Andriushchenko, and Bitterwolf(2019)}]{hein2019relu}
Hein, M.; Andriushchenko, M.; and Bitterwolf, J. 2019.
\newblock Why ReLU networks yield high-confidence predictions far away from the
  training data and how to mitigate the problem.
\newblock In \emph{Proceedings of the IEEE Conference on Computer Vision and
  Pattern Recognition}, 41--50.

\bibitem[{Hendrycks and Gimpel(2017)}]{hendrycks2017baseline}
Hendrycks, D.; and Gimpel, K. 2017.
\newblock A Baseline for Detecting Misclassified and Out-of-Distribution
  Examples in Neural Networks.
\newblock In \emph{International Conference on Learning Representations}.

\bibitem[{Hendrycks, Mazeika, and Dietterich(2019)}]{hendrycks2018deep}
Hendrycks, D.; Mazeika, M.; and Dietterich, T. 2019.
\newblock Deep Anomaly Detection with Outlier Exposure.
\newblock In \emph{International Conference on Learning Representations}.

\bibitem[{Hsu et~al.(2020)Hsu, Hong, Lee, and Liu}]{hsu2020query}
Hsu, Y.-C.; Hong, C.-Y.; Lee, M.-S.; and Liu, T.-L. 2020.
\newblock Query-Driven Multi-Instance Learning.
\newblock In \emph{Proceedings of the AAAI Conference on Artificial
  Intelligence}, volume~34.

\bibitem[{Kamath, Jia, and Liang(2020)}]{kamath2020selective}
Kamath, A.; Jia, R.; and Liang, P. 2020.
\newblock Selective question answering under domain shift.
\newblock \emph{arXiv preprint arXiv:2006.09462} .

\bibitem[{Karpathy and Fei-Fei(2015)}]{karpathy2015deep}
Karpathy, A.; and Fei-Fei, L. 2015.
\newblock Deep visual-semantic alignments for generating image descriptions.
\newblock In \emph{Proceedings of the IEEE conference on computer vision and
  pattern recognition}, 3128--3137.

\bibitem[{Kim, Jun, and Zhang(2018)}]{Kim2018}
Kim, J.-H.; Jun, J.; and Zhang, B.-T. 2018.
\newblock Bilinear Attention Networks.
\newblock In \emph{Advances in Neural Information Processing Systems 31},
  1571--1581.

\bibitem[{Kingma and Dhariwal(2018)}]{kingma2018glow}
Kingma, D.~P.; and Dhariwal, P. 2018.
\newblock Glow: Generative flow with invertible 1x1 convolutions.
\newblock In \emph{Advances in neural information processing systems},
  10215--10224.

\bibitem[{Lee et~al.(2018)Lee, Lee, Lee, and Shin}]{lee2018simple}
Lee, K.; Lee, K.; Lee, H.; and Shin, J. 2018.
\newblock A simple unified framework for detecting out-of-distribution samples
  and adversarial attacks.
\newblock In \emph{Advances in Neural Information Processing Systems},
  7167--7177.

\bibitem[{Liang, Li, and Srikant(2018)}]{liang2018enhancing}
Liang, S.; Li, Y.; and Srikant, R. 2018.
\newblock Enhancing The Reliability of Out-of-distribution Image Detection in
  Neural Networks.
\newblock In \emph{International Conference on Learning Representations}.

\bibitem[{Lin et~al.(2014)Lin, Maire, Belongie, Hays, Perona, Ramanan,
  Doll{\'a}r, and Zitnick}]{lin2014microsoft}
Lin, T.-Y.; Maire, M.; Belongie, S.; Hays, J.; Perona, P.; Ramanan, D.;
  Doll{\'a}r, P.; and Zitnick, C.~L. 2014.
\newblock Microsoft coco: Common objects in context.
\newblock In \emph{European conference on computer vision}, 740--755. Springer.

\bibitem[{Maaten and Hinton(2008)}]{maaten2008visualizing}
Maaten, L. v.~d.; and Hinton, G. 2008.
\newblock Visualizing data using t-SNE.
\newblock \emph{Journal of machine learning research} 9(Nov): 2579--2605.

\bibitem[{Mahendru et~al.(2017)Mahendru, Prabhu, Mohapatra, Batra, and
  Lee}]{mahendru2017promise}
Mahendru, A.; Prabhu, V.; Mohapatra, A.; Batra, D.; and Lee, S. 2017.
\newblock The Promise of Premise: Harnessing Question Premises in Visual
  Question Answering.
\newblock In \emph{Proceedings of the 2017 Conference on Empirical Methods in
  Natural Language Processing}, 926--935.

\bibitem[{Meinke and Hein(2019)}]{meinke2019towards}
Meinke, A.; and Hein, M. 2019.
\newblock Towards neural networks that provably know when they don't know.
\newblock In \emph{International Conference on Learning Representations}.

\bibitem[{Nguyen, Yosinski, and Clune(2015)}]{nguyen2015deep}
Nguyen, A.; Yosinski, J.; and Clune, J. 2015.
\newblock Deep neural networks are easily fooled: High confidence predictions
  for unrecognizable images.
\newblock In \emph{Proceedings of the IEEE conference on computer vision and
  pattern recognition}, 427--436.

\bibitem[{Pennington, Socher, and Manning(2014)}]{pennington2014glove}
Pennington, J.; Socher, R.; and Manning, C. 2014.
\newblock Glove: Global vectors for word representation.
\newblock In \emph{Proceedings of the 2014 conference on empirical methods in
  natural language processing (EMNLP)}, 1532--1543.

\bibitem[{Ray et~al.(2016)Ray, Christie, Bansal, Batra, and
  Parikh}]{ray2016question}
Ray, A.; Christie, G.; Bansal, M.; Batra, D.; and Parikh, D. 2016.
\newblock Question Relevance in VQA: Identifying Non-Visual And False-Premise
  Questions.
\newblock In \emph{Proceedings of the 2016 Conference on Empirical Methods in
  Natural Language Processing}, 919--924.

\bibitem[{Ren et~al.(2019)Ren, Liu, Fertig, Snoek, Poplin, Depristo, Dillon,
  and Lakshminarayanan}]{ren2019likelihood}
Ren, J.; Liu, P.~J.; Fertig, E.; Snoek, J.; Poplin, R.; Depristo, M.; Dillon,
  J.; and Lakshminarayanan, B. 2019.
\newblock Likelihood ratios for out-of-distribution detection.
\newblock In \emph{Advances in Neural Information Processing Systems},
  14680--14691.

\bibitem[{Ren et~al.(2015)Ren, He, Girshick, and Sun}]{ren2015faster}
Ren, S.; He, K.; Girshick, R.; and Sun, J. 2015.
\newblock Faster r-cnn: Towards real-time object detection with region proposal
  networks.
\newblock In \emph{Advances in neural information processing systems}, 91--99.

\bibitem[{Salimans et~al.(2017)Salimans, Karpathy, Chen, and
  Kingma}]{Salimans2017PixeCNN}
Salimans, T.; Karpathy, A.; Chen, X.; and Kingma, D.~P. 2017.
\newblock PixelCNN++: A PixelCNN Implementation with Discretized Logistic
  Mixture Likelihood and Other Modifications.
\newblock In \emph{ICLR}.

\bibitem[{Settles(2009)}]{settles2009active}
Settles, B. 2009.
\newblock Active learning literature survey.
\newblock Technical report, University of Wisconsin-Madison Department of
  Computer Sciences.

\bibitem[{Singh et~al.(2019)Singh, Natarjan, Shah, Jiang, Chen, Parikh, and
  Rohrbach}]{singh2019towards}
Singh, A.; Natarjan, V.; Shah, M.; Jiang, Y.; Chen, X.; Parikh, D.; and
  Rohrbach, M. 2019.
\newblock Towards VQA Models That Can Read.
\newblock In \emph{Proceedings of the IEEE Conference on Computer Vision and
  Pattern Recognition}, 8317--8326.

\bibitem[{Yu et~al.(2019)Yu, Yu, Cui, Tao, and Tian}]{yu2019deep}
Yu, Z.; Yu, J.; Cui, Y.; Tao, D.; and Tian, Q. 2019.
\newblock Deep modular co-attention networks for visual question answering.
\newblock In \emph{Proceedings of the IEEE Conference on Computer Vision and
  Pattern Recognition}, 6281--6290.

\bibitem[{Yu et~al.(2017)Yu, Yu, Fan, and Tao}]{yu2017multi}
Yu, Z.; Yu, J.; Fan, J.; and Tao, D. 2017.
\newblock Multi-modal factorized bilinear pooling with co-attention learning
  for visual question answering.
\newblock In \emph{Proceedings of the IEEE international conference on computer
  vision}, 1821--1830.

\bibitem[{Yu et~al.(2018)Yu, Yu, Xiang, Fan, and Tao}]{yu2018beyond}
Yu, Z.; Yu, J.; Xiang, C.; Fan, J.; and Tao, D. 2018.
\newblock Beyond bilinear: Generalized multimodal factorized high-order pooling
  for visual question answering.
\newblock \emph{IEEE transactions on neural networks and learning systems}
  29(12): 5947--5959.

\bibitem[{Zhang et~al.(2018)Zhang, Zhao, Li, and Zong}]{zhang2018attention}
Zhang, J.; Zhao, Y.; Li, H.; and Zong, C. 2018.
\newblock Attention with sparsity regularization for neural machine translation
  and summarization.
\newblock \emph{IEEE/ACM Transactions on Audio, Speech, and Language
  Processing} 27(3): 507--518.

\end{thebibliography}
}
\appendix
\onecolumn
\section{A.Visualization of Joint Feature Vectors}
VQA models are known to have a bias on the questions than the images \cite{goyal2017making}.
In this study, we argue that if a VQA model has more bias on one modality, it is hard to detect anomalies of the other modality because the abnormal source vanishes after multimodal feature fusion.
When the joint features of anomalies cannot be distinguished from normal samples, the confidence of the answer is not the best way to detect various anomalies in VQA. 

\begin{figure}[h]
\centering
\includegraphics[width=8cm]{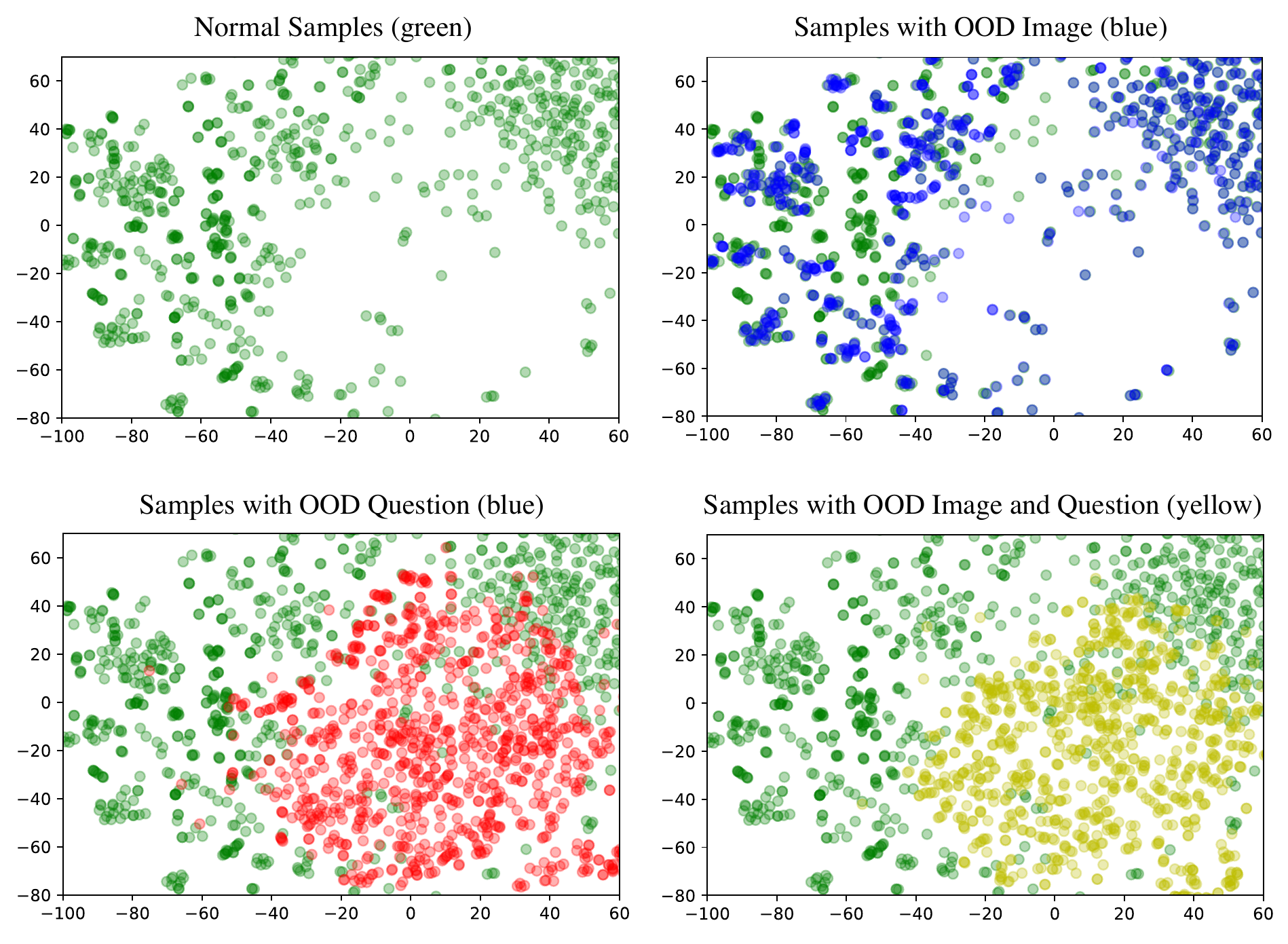}
\caption{T-SNE visualization of joint features: normal samples (green) and samples with out-of-distribution image (TinyImageNet, blue), question (IMDB, red), and both image and question (TinyImageNet and IMDB, yellow)}
\label{fig:feature_viz}
\end{figure}

We visualize the joint features of normal and abnormal samples by T-SNE \cite{maaten2008visualizing} (Fig.~\ref{fig:feature_viz}).
We use the BUTD model and extract the joint features that integrate image and question features by element-wise multiplication \cite{anderson2018bottom}.
If a sample has an out-of-distribution question (red and yellow), the vectors of joint features are distinguishable from normal samples (green).
However, we cannot distinguish the joint features of abnormal samples, when the samples have an out-of-distribution image and in-distribution question (blue).
The results imply that the abnormality in input images can vanish after multimodal feature fusion because of the bias on the question input.

\section{B. Optimality for Regularization of Attention Networks}
In this section, we show that the optimal solution of our regularizing attention networks is uniform distribution.
Without loss of generality, we prove theorem 1. 
\vspace{4pt}

\textbf{Theorem 1} Suppose $\mathbf{x} = (x_1, ..., x_K)$ where $0 \le x_i\le 1$ for $i=1,...,K$ and $\sum_{i=1}^{K}x_i = 1$.
Consider the problem to maximize $f(\mathbf{x})=\sum_{i=1}^{K}{\log (1-x_i)}$.
Then, $\mathbf{x}^{*}$ is the optimal solution of this problem if and only if $x_i^{*}=\frac{1}{K}$ for $i=1,...,K$.

\begin{proof}
We use Lagrangian multiplier $\lambda$ to solve above constrained optimization problem.
Suppose $g(\mathbf{x})=\sum_{i=1}^{K}{x_i}$.
Then, the optimal solution solves the following equations.
\begin{align}
    \nabla f(\mathbf{x}^*) &= \lambda \nabla g(\mathbf{x}^*) \label{s_eq:1} \\ 
    g(\mathbf{x}^*) &= 1 \label{s_eq:2}
\end{align}
Considering $ \nabla f(\mathbf{x}) = (\frac{1}{x_1 - 1}, ..., \frac{1}{x_K - 1})$ and $\nabla g(\mathbf{x})= (1, ..., 1)$, the optimal solution $\mathbf{x}^*$ is an uniform distribution such that 

\begin{equation}
    \mathbf{x}^* = (1+\frac{1}{\lambda}, ..., 1+\frac{1}{\lambda})
\end{equation}
where $\lambda \neq 0$.
Then, we get $\lambda=\frac{K}{1-K}$ and $\mathbf{x}^* = (\frac{1}{K},...,\frac{1}{K})$ to satisfy Eq~(\ref{s_eq:2}).
\end{proof}
Thus, our regularization makes a VQA model have the uniform attention weight distribution on abnormal samples.

\section{C. Additional Experimental Results}
\subsection{Experimental Setup}
$K=36$ objects are detected by pretrained faster R-CNN \cite{ren2015faster}, and a 2048 dimensional vector for each object is extracted by pretrained ResNet-152 \cite{he2016deep}.
Question tokens are trimmed to a maximum of 14 words, and pretrained GloVe \cite{pennington2014glove} is used for word embedding.
The batch size is 256.
For BAN, the dimension of fully connected layer is 1280, and, the size of glimpse is two.
Other models use 1024 dimensional vector for the fully connected layer.
The initial learning rate is 0.002 and all trainable parameters are updated by Adamax optimizer in Pytorch.
The parameters are updated in 30 epochs, and we use the best model to evaluate the robustness of the model to various anomalies.
We use two V100 GPUs for training VQA models with pretrained features, and the training is finished in 4 hours.
Other hyperparameters are the same as those in the original paper for each model.

For the regularization of the attention network, we use training samples of TinyImage, VNQ, and QRPE for $P_{\text{anomaly}}$ in Eq~(5).
Each mini-batch for regularization consists of balanced samples, which equally contain TinyImage, VQA, and QRPE samples.
In addition, Note that there is no overlap of anomaly data between data for training and evaluation.
We fine-tune the pretrained VQA models in 15 epochs, and the $\lambda$ in Eq~(5) is set to 0.00001.

To compute the maximum attention score, we average the maximum attention scores of all heads in the attention network to consider all information in the multiple attention heads.
Some VQA models, such as MCAN, have multiple layers, and each layer has an attention module to extract hidden features for the input of the following layer.
Then, we use the first attention layer because it focuses on the most relevant information and filters out irrelevant information in a pair of images and questions, and the later layers repeatedly elaborate on the hidden features of the image and question.
All codes are implemented using Pytorch 0.4.1 and are available in public.

\subsection{Results of Task 3: Out-of-distribution Image and Question}
The experimental results of task 3 (out-of-distribution image and question) is in Table~\ref{tab:vq_ood_overall}.
All combinations of out-of-distribution images (MNIST, SVHN, FashionMNIST, CIFAR10, and TinyImageNet) and questions (20 Newsgroup, Reuter52, and IMDB) are considered in task 3.
This situation when both image and question are from out-of-distribution is an extreme case, but we contain task 3 to evaluate the robustness of VQA models on the extreme case that is expected to be detected more easily.

The results show that the attention-based approach and our regularization outperform compared methods.
For all VQA models and datasets, attention-based approach shows almost perfect detection of out-of-distribution samples.
When we consider that task 3 is rather extreme and easy than task 1 and 2, the superior results are acceptable.

\begin{table*}
\centering
\caption{Task 3 performances: out-of-distribution image \& question detection AUROCs of VQA models.}
\begin{tabular}{l|c|c|c}
\hline
AUROC		 &	BUTD	&	MHB+ATT	&	BAN  \\ \hline
20 Newsgroup& \multicolumn{3}{c}{MSP/MSP(T)/OE-MSP(T)/MAP(T)/RA-MAP(T)} \\ \hline 
MNIST	&		77.9/94.3/71.5/\textbf{100}/99.9     & 49.8/38.1/\textbf{97.0}/96.9/96.7  & 51.9/52.5/57.1/99.9/\textbf{100} \\
SVHN	&		77.6/94.0/70.5/99.9/\textbf{99.9}    & 49.8/38.1/96.8/\textbf{97.2}/96.3  & 51.8/52.1/56.1/99.8/\textbf{100}	\\
FashionMNIST	&		77.8/94.4/71.8/\textbf{100}/99.9 &   49.7/38.0/\textbf{97.3}/97.1/97.2    &   51.9/52.4/57.5/99.9/\textbf{100}	\\
CIFAR10	&		77.9/94.1/69.4/99.9/\textbf{99.9} & 49.7/38.2/96.4/\textbf{97.0}/95.6 & 51.7/52.2/54.9/99.8/\textbf{100}	\\ 
TinyImageNet	&		76.6/93.3/73.8/100/\textbf{100}  &   49.5/37.6/96.9/\textbf{97.6}/95.6    &   51.9/52.0/61.8/100/\textbf{100}	\\ \hline
Reuters52& \multicolumn{3}{c}{MSP/MSP(T)/OE-MSP(T)/MAP(T)/RA-MAP(T)} \\ \hline 
MNIST	&		81.5/95.8/73.6/99.8/\textbf{99.9} & 47.3/33.8/\textbf{97.7}/96.6/97.3 &   51.4/52.8/56.5/99.9/\textbf{100}	\\
SVHN	&		81.2/95.5/73.0/\textbf{100}/99.9 &   47.3/33.8/\textbf{97.5}/96.8/97.0 &  51.4/52.6/56.0/99.8/\textbf{100}	\\
FashionMNIST	&		81.4/95.8/73.9/\textbf{100}/99.9 & 47.3/33.8/\textbf{98.0}/96.8/97.7  &   51.5/52.9/57.2/99.9/\textbf{100}	\\
CIFAR10	&		81.6/95.7/72.2/\textbf{100}/99.9 &   47.4/34.2/\textbf{97.3}/96.6/96.4    &   51.3/52.5/54.5/99.8/\textbf{100}	\\ 
TinyImageNet	&		80.0/94.8/75.2/100/\textbf{100} & 47.1/33.5/97.7/97.3/\textbf{98.9}   &   51.4/52.3/61.5/100/\textbf{100}	\\ \hline
IMDB& \multicolumn{3}{c}{MSP/MSP(T)/OE-MSP(T)/MAP(T)/RA-MAP(T)} \\ \hline 
MNIST	&		73.7/92.9/69.7/\textbf{100}/99.9 &   44.4/30.9/96.1/\textbf{97.1}/96.6    &   44.1/40.2/57.6/99.7/\textbf{100}	\\
SVHN	&		73.5/92.5/68.4/\textbf{100}/99.9 &   44.3/30.8/95.9/\textbf{97.4}/96.2    &   44.0/40.0/57.2/99.8/\textbf{100}	\\
FashionMNIST	&		73.7/93.0/70.2/\textbf{100}/99.9 &   44.2/30.7/96.5/\textbf{97.2}/95.4    &   43.9/39.8/55.7/99.7/\textbf{100}	\\
CIFAR10	&		73.8/92.7/67.3/\textbf{100}/99.9 &   44.4/31.0/95.5/\textbf{97.3}/95.4    &   43.9/39.8/55.7/99.7/\textbf{100}	\\ 
TinyImageNet	&		72.2/91.5/72.4/100/\textbf{100} & 43.9/30.3/96.0/97.7/\textbf{98.8} & 44.1/39.7/62.3/100/\textbf{100}	\\ \hline
\end{tabular}
\label{tab:vq_ood_overall}
\end{table*}

\subsection{Comparison with Minimization of Attention Variance}
Another alternative to make an attention weight uniform is minimization of variance of attention distribution.
When the variance of attention weights is minimized, the variance is zero and the all attention weights are same.

In Table~\ref{tab:acc_var_log}, we find that minimizing the variance of attention networks (RA-VAR) induces more degradation of VQA accuracy than maximization of our regularization $\sum_{i=1}^{K}\sum_{j=1}^{M}\log(1-A_{ij})$ (RA-LOG).
In this study, we prefer to use RA-LOG because anomaly detection in VQA is useless if the VQA accuracy does not conserved.

Although the accuracy drop, RA-VAR is also effective to detect various abnormal samples in test time (Table~\ref{tab:vq_ood_overall}).
The results imply that our assumption of attention networks appropriates for anomaly detection in VQA regardless of its implementation.


\begin{table*}
\centering
\caption{Degradation of accuracy by additional training for anomaly detection}
\begin{tabular}{l|cccc}
\hline
Accuracy (\%)     & Base & OE &  RA-VAR & RA-LOG \\ \hline 
BUTD	&	62.55 & 54.87(-7.68) & 61.56(-0.99)& 61.91(-0.64)\\
MHB+ATT	&	63.33 & 62.38(-0.95) & 62.54(-0.79)& 62.77(-0.56) \\
BAN	&	63.81 & 61.92(-1.89) & 63.37(-0.44)& 63.74(-0.07) \\\hline
\end{tabular}
\label{tab:acc_var_log}
\end{table*}

\begin{table*}
\centering
\caption{Comparison of regularization implementations. RA-VAR: minimization of variance of attention networks. RA-LOG: original implementation in the paper: maximization of $\sum_{i=1}^{K}\sum_{j=1}^{M}\log(1-A_{ij})$}
\scalebox{0.9}{
\begin{tabular}{l|c|c|c}
\hline
AUROC		 &	BUTD	&	MHB+ATT	&	BAN \\ \hline
Image& \multicolumn{3}{c}{MAP(T)/RA-VAR/RA-LOG} \\ \hline 
MNIST	&		89.0/97.4/\textbf{97.8}	&	89.9/93.2/\textbf{94.7}	&	99.0/92.2/\textbf{100}	\\
SVHN	&		90.3/97.5/\textbf{97.9}	&	89.7/95.4/\textbf{96.2}	&	100/93.4/\textbf{100}	\\
FashionMNIST	&		89.6/97.4/\textbf{97.8}	&	90.5/94.6/\textbf{95.7}	&	99.9/93.0/\textbf{100}	\\
CIFAR10	&		90.5/97.5/\textbf{98.0}	&	89.9/96.3/\textbf{96.9}	&	100/94.1/\textbf{100}	\\ 
TinyImageNet	&		92.7/98.5/\textbf{99.7}	&	91.5/99.1/\textbf{99.2}	&	100/96.0/\textbf{100}	\\ \hline
Question& \multicolumn{3}{c}{MAP(T)/RA-VAR/RA-LOG} \\ \hline 
20 Newsgroup	&		78.2/\textbf{97.1}/95.5	&	78.9/\textbf{97.4}/92.6	&	81.7/\textbf{93.2}/87.3	\\
Reuters52	&		76.4/\textbf{97.7}/97.0	&	77.4/\textbf{98.6}/94.3	&	81.7/\textbf{93.9}/87.3	\\
IMDB	&		78.2/\textbf{95.1}/92.8	&	77.9/\textbf{95.9}/91.1	&	78.1/\textbf{89.4}/82.5	\\ \hline
Irrelevant Q& \multicolumn{3}{c}{MAP(T)/RA-VAR/RA-LOG} \\ \hline 
VNQ	&		60.9/88.7/\textbf{98.3}	&	75.8/\textbf{99.7}/99.6	&	73.5/\textbf{96.8}/90.0	\\
QRPE	&		49.6/\textbf{89.0}/84.8	&	56.6/\textbf{94.4}/94.0	&	61.7/\textbf{79.5}/63.6	\\ \hline
\end{tabular}}
\label{tab:var_log_comp}
\end{table*}

\begin{figure}
\centering
\includegraphics[width=14cm]{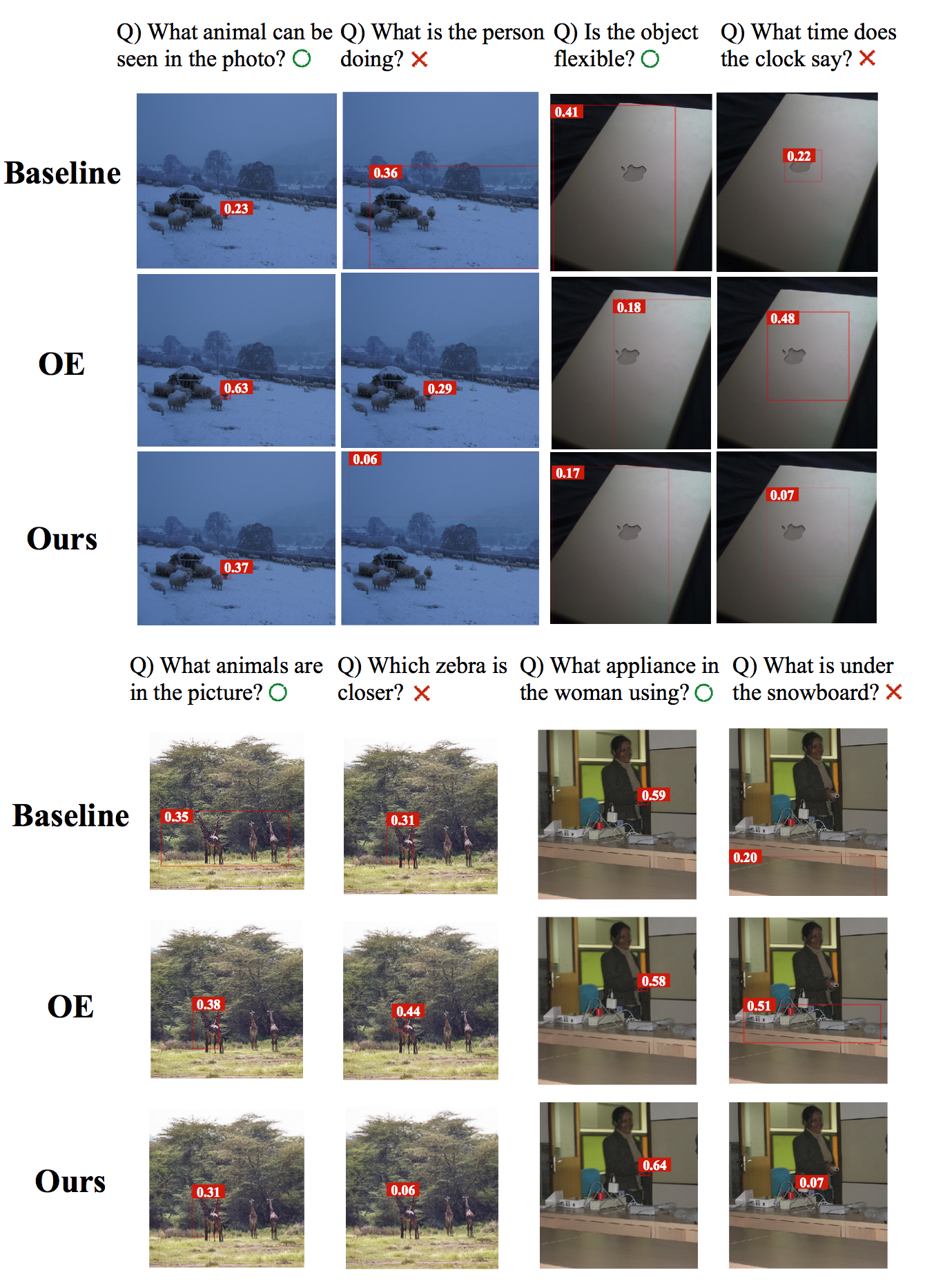}
\caption{Results of top-1 attention on samples with irrelevant question (1). Baseline (BUTD), baseline + OE, and baseline + our regularization are compared.}
\label{fig:comp_irr_1}
\end{figure}

\begin{figure}
\centering
\includegraphics[width=14cm]{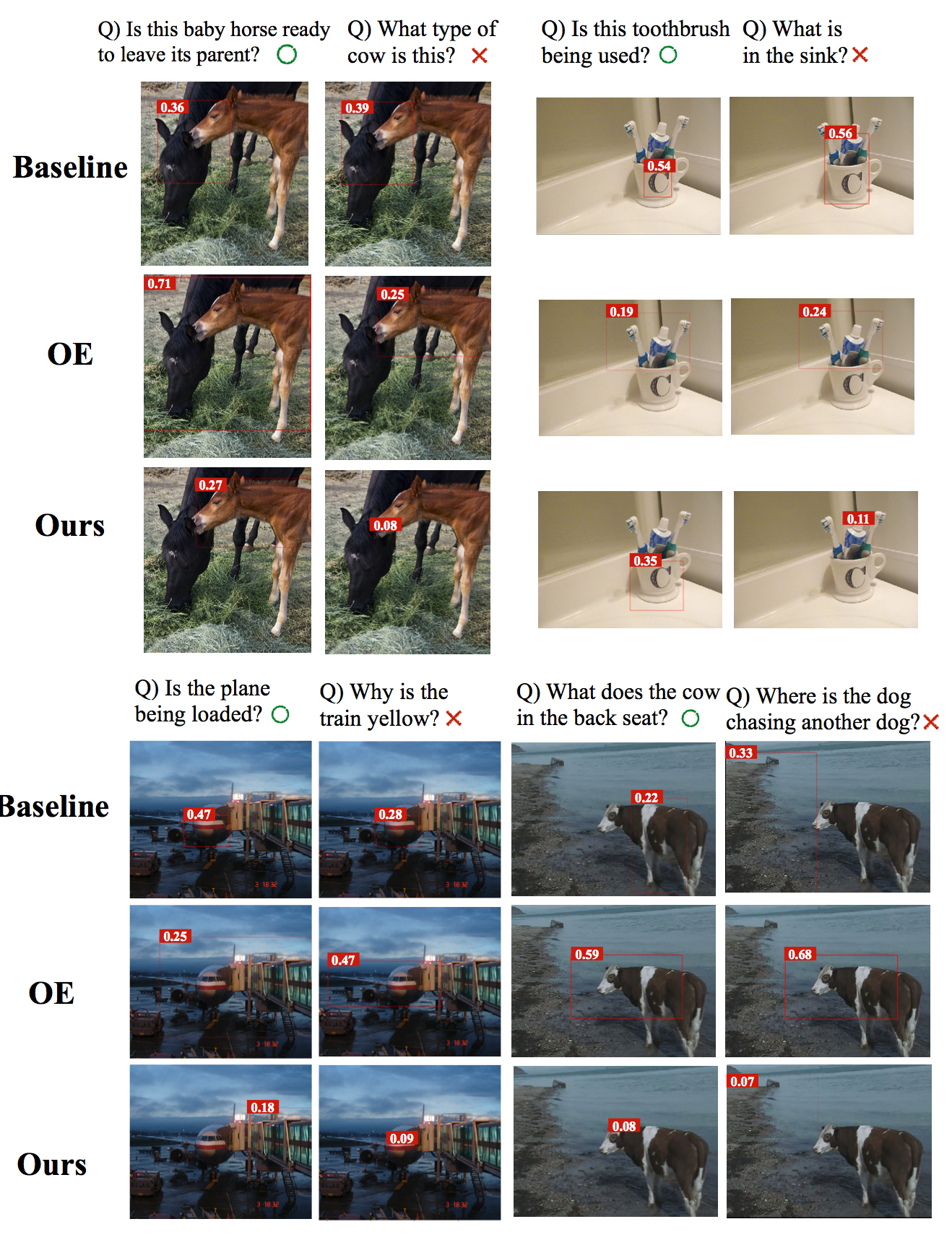}
\caption{Results of top-1 attention on samples with irrelevant question (2). Baseline (BUTD), baseline + OE, and baseline + our regularization are compared.}
\label{fig:comp_irr_2}
\end{figure}

\begin{figure}
\centering
\includegraphics[width=14cm]{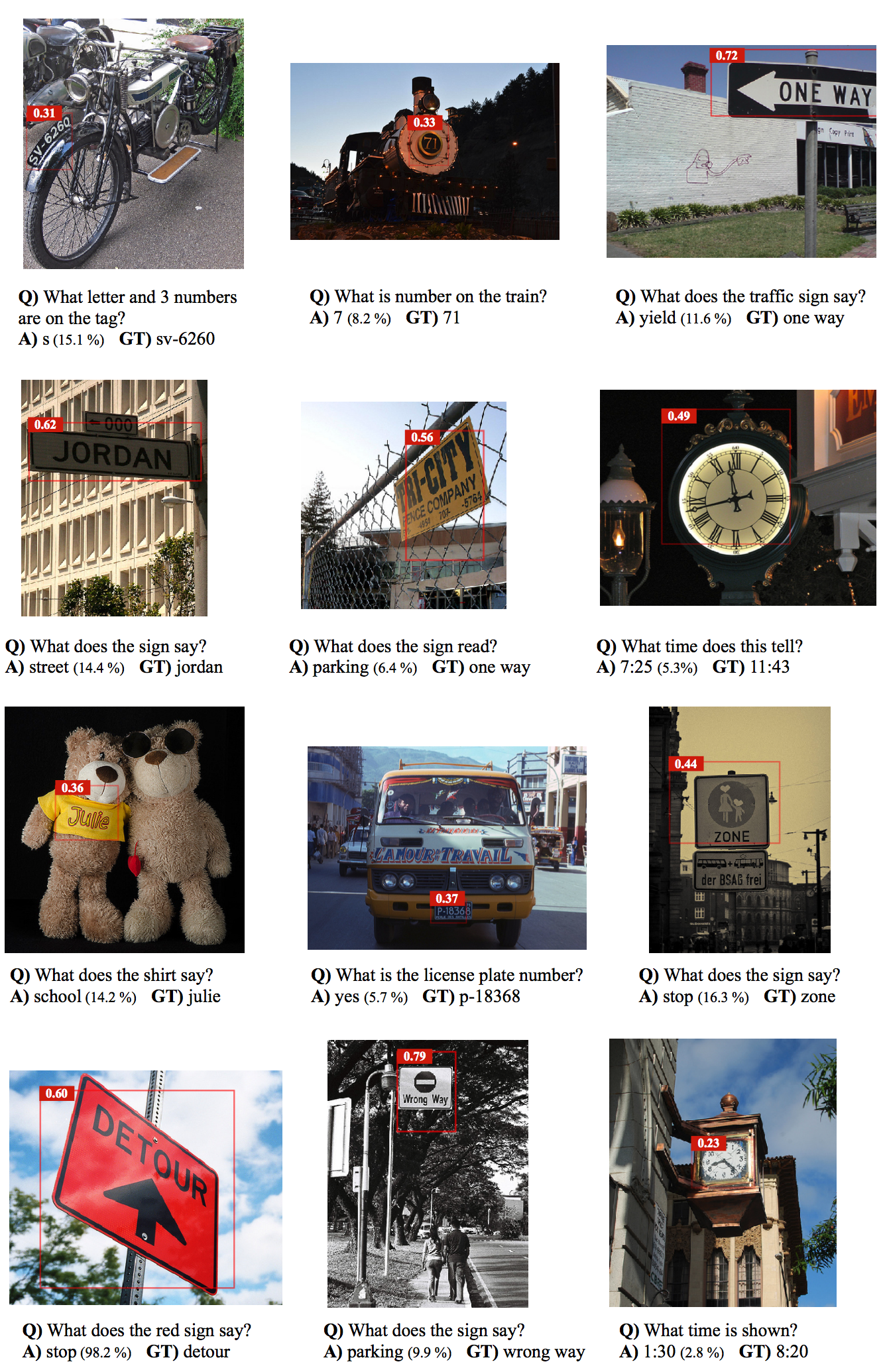}
\caption{Results of top-1 attention on samples with undefined answer (reading texts and numbers). Baseline (BUTD) makes a proper attention, but results in completely different output.}
\label{fig:comp_undef}
\end{figure}

\section{D. Qualitative Analysis of Unanswerable Samples}
\subsection{Examples of Task 4: Irrelevant Question}
Experimental results of MAP (T) imply that attention networks in VQA have a bias of highly confident attention to a prominent object regardless of the question relevance. 
In Fig.~\ref{fig:comp_irr_1} and ~\ref{fig:comp_irr_2}, We attach examples of the results of maximum attention scores on samples with irrelevant questions before and after our regularization.
Considering the sum of the attention scores is 1.0 over 36 objects (each object has about 0.03 in a uniform distribution), baseline models without regularization or with outlier exposure \cite{hendrycks2018deep} still give strong attention on a prominent visual object, even though the questions are irrelevant and there is not related visual object in the images.
However, after regularization (ours), the attention networks avoid giving strong attention to any prominent object when the question is irrelevant to the image, and the attention scores become close to uniform.
Of course, if the question is relevant to the image, our model makes proper and strong attention to the visual object.

\subsection{Examples of Task 5: Undefined Answer}
We further study the samples with undefined answers and attached various examples in Fig.~\ref{fig:comp_undef}.
We find that about 40\% (1676/4303) of samples with undefined answers require to read texts or numbers in input images.
Note that the VQA model makes the right attention on the samples with undefined answers, but makes strange outputs because of their unanswerability by predefined answer candidates.
It is a common limitation of recent VQA models, which solve VQA as a prediction problem, such as classification or regression, because there are infinite combinations of text and numbers in the real-world \cite{singh2019towards}.

\begin{table}
\centering
\caption{Anomaly Detection Performances (task 1, 2, and 4) of regression-based VQA model (NTN).}
\begin{tabular}{l|c}
\hline
AUROC		 &	BUTD + NTN \\ \hline 
Image & MSP/MSP(T)/OE-MSP(T)/MAP(T)/RA-MAP(T) \\ \hline 
MNIST	&		67.5/84.3/64.3/\textbf{93.8}/92.3\\
SVHN	&		67.7/85.5/66.0/\textbf{94.5}/92.9\\
FashionMNIST	&		67.6/85.1/66.3/\textbf{94.3}/92.5\\
CIFAR10	&		67.8/85.9/66.9/\textbf{94.6}/93.1\\
Tiny & 67.6/86.7/74.9/\textbf{95.8}/94.1 \\ \hline
Question & MSP/MSP(T)/OE-MSP(T)/MAP(T)/RA-MAP(T) \\ \hline 
20 Newsgroup	&	68.9/83.4/70.3/78.1/\textbf{92.7} \\
Reuters52	&	69.9/83.5/68.4/77.4/\textbf{94.0} \\
IMDB	&	63.3/80.6/66.9/77.5/\textbf{89.1} \\ \hline
Irrelevant Q & MSP/MSP(T)/OE-MSP(T)/MAP(T)/RA-MAP(T) \\ \hline 
VNQ	&		83.8/92.0/76.4/69.9/\textbf{96.0} / \\
QRPE	&		67.3/70.8/79.8/45.4/\textbf{89.1} \\
\hline
\end{tabular}
\label{tab:ntn_result}
\end{table}

\section{E. Anomaly Detection of Regression-based VQA Models}
Although classification-based VQA models show promising results on the VQA v2 challenge \cite{anderson2018bottom,Kim2018,yu2019deep}, regression-based VQA models are also feasible and important approaches in VQA studies.
The main differences between classification- and regression-based approaches are that regression-based VQA models take an answer candidate as the input together, compute likelihood $p(\mathbf{a},\mathbf{v},\mathbf{q})$ scores of all answer answer candidates, and select the most likely answer candidate as the prediction:
\begin{gather}
\theta^{*}=\underset{\theta}{\operatorname{argmax}}\mathbb{E}_{p_{\mathcal{D}}}\left[\log p_{\theta}\left(\mathbf{a}, \mathbf{v}, \mathbf{q}\right)\right], \\
S({\mathbf{v}, \mathbf{q}} ; T) = \max_i{ p_{\theta}(\mathbf{a}_i,\mathbf{v}, \mathbf{q} ; T)}.    
\end{gather}

One can speculate that regression-based VQA models, which jointly consider an image, a question, and the answer, are free of anomaly detection.
Thus, we evaluate the robustness of regression-based anomaly detection methods.
We modify the model architecture of BUTD based on common approaches, neural tensor networks (NTN) \cite{bai2018deep}.
Note that although regression-based VQA models compute a likelihood on every answer candidate, cross-modal attention networks are located before the multi-modal feature fusion, and our attention-based anomaly detection can still be used (see the details in \cite{bai2018deep}).
The results in Table~\ref{tab:ntn_result} imply that the regression-based approach still suffers from various anomalies, and attention-based anomaly detection makes the VQA model robust to them.

\end{document}